\documentclass{arxiv_preprint}

\usepackage{amsmath,amsfonts,bm}

\def\eqref#1{equation~\ref{#1}}
\def\Eqref#1{Equation~\ref{#1}}

\def\1{\bm{1}}

\DeclareMathAlphabet{\mathsfit}{\encodingdefault}{\sfdefault}{m}{sl}
\SetMathAlphabet{\mathsfit}{bold}{\encodingdefault}{\sfdefault}{bx}{n}

\usepackage{graphicx}
\usepackage{colortbl}
\usepackage{amssymb}
\usepackage{multirow}
\usepackage{hyperref}
\usepackage{url}
\usepackage{booktabs}

\title{\textcolor{metablue}{EvoMem-VLA}: State-Evolution Memory for Long-Horizon Robot Manipulation}
\author[1,2,4,\ddagger]{Yuheng Na}
\author[2,\ddagger,\dagger]{Zhide Zhong}
\author[2,\ddagger]{Junjie He}
\author[2,\ddagger]{Junfeng Li}
\author[2]{Haodong Yan}
\author[2]{Jiaan Wang}
\author[2]{Jiaguan Zhu}
\author[2]{Yangyang Zheng}
\author[3]{Tianyu Huang}
\author[2,\ast]{Haoang Li}
\affiliation[1]{Xi'an Jiaotong University}
\affiliation[2]{The Hong Kong University of Science and Technology (Guangzhou)}
\affiliation[3]{The Chinese University of Hong Kong}
\affiliation[4]{Xbotics Embodied AI Community}
\contribution[\ddagger]{Equal Contribution}
\contribution[\dagger]{Project lead}
\contribution[\ast]{Corresponding author.}
\hypersetup{
  pdftitle={EvoMem-VLA: State-Evolution Memory for Long-Horizon Robot Manipulation},
  pdfauthor={Yuheng Na, Zhide Zhong, Junjie He, Junfeng Li, Haodong Yan, Jiaan Wang, Jiaquan Zhu, Yangyang Zheng, Tianyu Huang, Haoang Li}
}

\abstract{
Most vision-language-action (VLA) models rely on current observations and lose task-relevant evidence once it leaves view, limiting performance on long-horizon, memory-dependent tasks. Existing efforts incorporate compressed historical features or sparse visual keyframes. However, isolated snapshots can leave the policy uncertain about what changed during past interactions and which action should follow. To overcome this limitation, we propose EvoMem-VLA, which constructs state-evolution memory by explicitly encoding and retaining observed changes between historical states. These change representations preserve evidence of interaction outcomes, allowing the policy to track task progress beyond isolated snapshots. Specifically, we introduce conditional delta tokenization to encode ordered frame pairs into directional, source-conditioned delta tokens, each associated with its corresponding state evidence. A shared VLM backbone supports task-adaptive routing: normal long-horizon tasks follow a direct action route, whereas multi-stage tasks use a subtask route that generates an executable subtask as an additional input for action generation. With a single jointly trained policy for each simulation benchmark, EvoMem-VLA achieves success rates of 80.7\% on RMBench, 82.0\% on RoboMME and 83.8\% across four real-world tasks spanning two robot embodiments. These results represent substantial improvements over the previous state of the art in all three evaluation settings.
\par\smallskip
\noindent\textbf{Project Page:} \url{https://yuhengna.github.io/EvoMem-VLA/}
}

\begin{document}
\maketitle

\section{Introduction}
\label{sec:introduction}

Vision-language-action (VLA) models transfer pretrained visual and semantic knowledge to robot control, enabling increasingly general instruction following and manipulation across tasks~\citep{rt2,openvla,pi0,xvla}. However, their performance remains limited on long-horizon, memory-dependent tasks. This is because a policy conditioned only on the current observation and instruction lacks access to past events that determine future actions~\citep{hamlet,memoryvla,eventvla}. As a result, the policy may lose track of task progress and generate an incorrect action.

\begin{figure}[t]
\centering
\includegraphics[width=\textwidth]{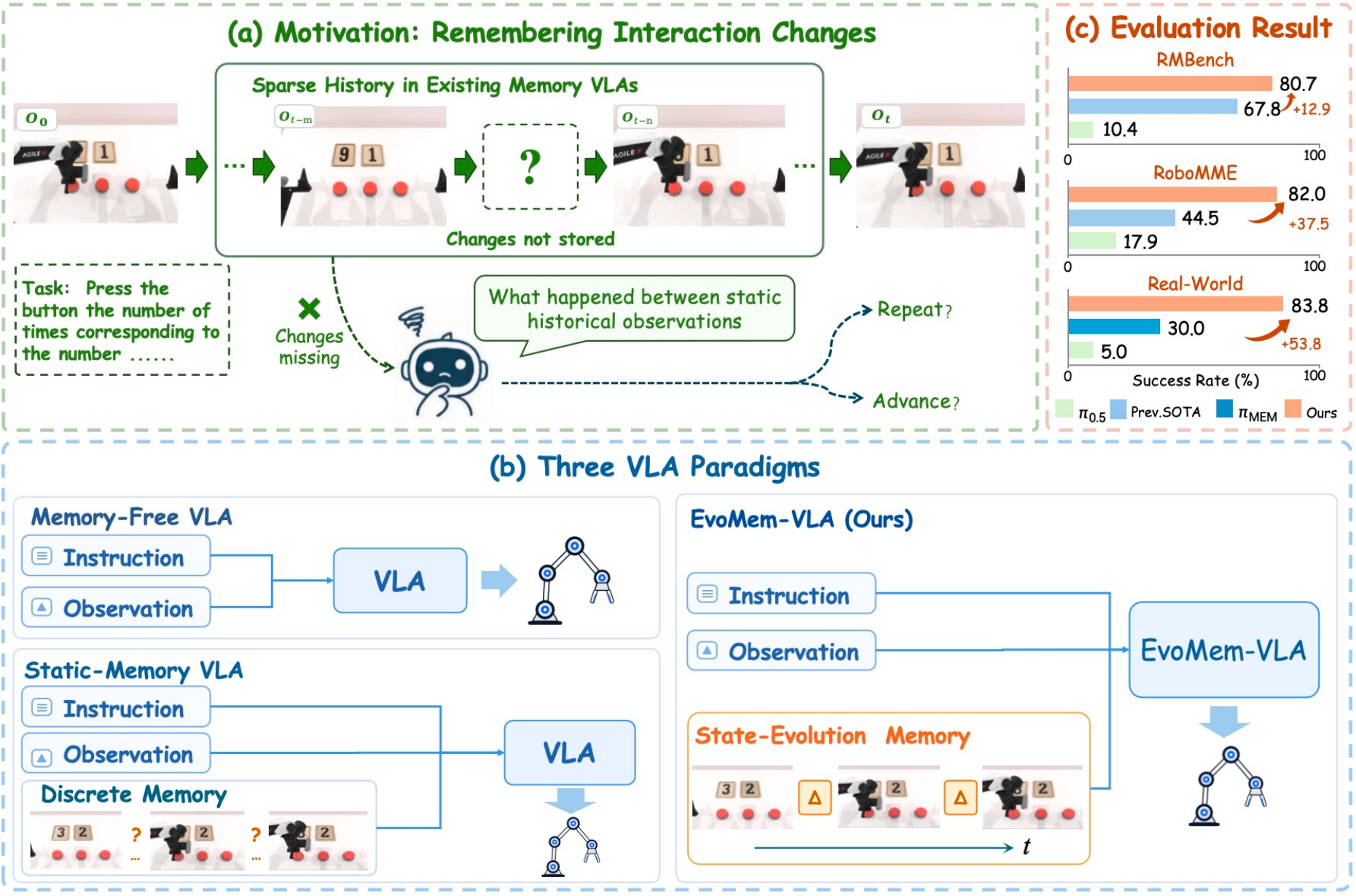}
\caption{\textbf{Overview of EvoMem-VLA.} EvoMem-VLA uses conditional delta tokenization to retain observed changes in state-evolution memory. (a) Motivation: sparse historical observations can leave important interaction changes unrecorded. (b) Comparison of memory-free VLAs, static-memory VLAs, and EvoMem-VLA. (c) State-of-the-art results on RMBench, RoboMME, and real-world tasks.}
\label{fig:evomem_overview}
\end{figure}
Recently, there have been efforts to augment VLAs with historical memory. Existing approaches represent history using discrete state features~\citep{hamlet,memoryvla,roboflamingo} or sparse visual keyframes~\citep{eventvla,kcvla,memer}. However, task-relevant changes often remain implicit in the stored history (Figure~\ref{fig:evomem_overview}(a)). Long-horizon control requires the policy to relate remembered states to changes during task execution. Without an explicit representation of these changes, the policy must extract them from sparse state evidence when determining task progress and selecting the next action. This motivates a memory representation that retains both what was observed and how it changed relative to an earlier observation.

To address this limitation, we propose EvoMem-VLA, a state-evolution memory paradigm that explicitly encodes and retains observed interaction changes between sparse historical observations (Figure~\ref{fig:evomem_overview}(b)). By linking change representations to corresponding visual states, this memory preserves evidence of interaction changes. Therefore, compared with VLAs that rely on discrete state memory, our policy can distinguish interaction processes that discrete history alone leaves ambiguous and use explicit change evidence to generate the appropriate next action. We instantiate this paradigm as a unified VLA with multi-scale state-evolution memory and task-adaptive routing. Specifically, to explicitly represent the visual change from the source frame to the target frame, we introduce conditional delta tokenization~\citep{deltatok} to encode ordered frame pairs sampled from recent observations and selected keyframes into delta tokens. We then construct state-evolution memory by arranging these delta tokens alongside the initial frame, recent historical frames, and selected keyframes~\citep{eventvla} in temporal order. Next, the state-evolution memory is fed into a shared VLM backbone together with the task instruction and current observation. Additionally, the VLM backbone supports task-adaptive routing: for normal long-horizon tasks without multi-stage planning, it predicts actions directly; for multi-stage tasks, it first generates an executable subtask and then appends it to the original input context to guide action generation~\citep{wla0}.

Our contributions are summarized as follows:

\begin{enumerate}
    \item We propose EvoMem-VLA, a state-evolution memory paradigm that explicitly retains observed interaction changes alongside historical states. This enables policies to use remembered change evidence for action generation rather than infer what happened from sparse historical states alone.

    \item We develop a unified VLA that organizes conditional delta tokens with initial, recent, and keyframe observations into multi-scale state-evolution memory. Its shared VLM backbone supports task-adaptive routing for direct action prediction or action generation conditioned on an executable subtask.

    \item EvoMem-VLA achieves state-of-the-art success rates of 80.7\% on RMBench~\citep{rmbench}, 82.0\% on RoboMME~\citep{robomme}, and 83.8\% across four real-world tasks on two robot embodiments. We use one jointly trained policy per simulation benchmark, with RMBench ablations further validating the contribution of conditional delta tokenization.
\end{enumerate}

\section{Related Work}
\label{sec:related_work}

\paragraph{Vision-Language-Action Models.}
Vision-language-action (VLA) models adapt pretrained vision-language representations to language-conditioned robot control~\citep{spatialvla,smolvla}. RT-2 and OpenVLA formulate action generation as autoregressive prediction over discretized action tokens~\citep{rt2,openvla}. More recent approaches introduce continuous or hybrid action generation~\citep{openvlaoft}: $\pi_0$ uses a flow-matching action expert~\citep{pi0}, CogACT and DexVLA condition diffusion-based action experts on VLM features~\citep{cogact,dexvla}, and HybridVLA combines autoregressive prediction with diffusion action generation~\citep{hybridvla}. X-VLA studies policy scaling across heterogeneous robot embodiments~\citep{xvla}. Despite advances in action representation and policy scaling, these models do not inherently maintain a persistent record of task execution. When earlier evidence leaves the observation window, a policy may lose track of what it has observed or completed, even when this history is needed to disambiguate visually similar states requiring different actions.

\paragraph{Memory Augmented Vision-Language-Action Models.}
History-aware policies address this limitation by retaining or transforming past observations~\citep{fourdvla,swiftvla,sam2act,mem}. RoboFlamingo compresses temporal vision-language features into an LSTM policy state~\citep{roboflamingo}, while TraceVLA overlays tracked point trajectories on current images as visual motion prompts~\citep{tracevla}. HAMLET encodes observations as temporally discriminative moment tokens and aggregates them through a lightweight causal memory module~\citep{hamlet}. MemoryVLA maintains perceptual and cognitive representations in long-term memory and retrieves relevant entries for action prediction~\citep{memoryvla}. EventVLA retains the initial observation and a recent history window alongside event-triggered keyframes in bounded memory~\citep{eventvla}. MemER uses a high-level policy to select and consolidate keyframes, predict the current language subtask, and delegate execution to a separate low-level controller~\citep{memer}. Complementary approaches use language and reasoning to guide action generation~\citep{ecot,rth,thinkact}: on multi-subtask data, $\pi_{0.5}$ predicts a semantic subtask before generating the corresponding action chunk~\citep{pi05}, while WLA-0 jointly learns textual-subtask prediction, future-state prediction, and action synthesis~\citep{wla0}. These methods provide remembered states, motion cues, and subtask context through distinct interfaces. However, the cited state-oriented memories lack dedicated inter-state change representations, while TraceVLA's short-window motion traces are not integrated with sparse long-term state evidence. MemER provides history to its high-level policy, but its controller receives only the predicted subtask, current images, and robot state. In contrast, EvoMem-VLA organizes multi-scale visual states and learned inter-state changes in state-evolution memory shared by subtask and action prediction. Tasks without textual-subtask supervision retain a direct action route through the same VLA policy.

\section{Method}
\label{sec:method}

\begin{figure*}[t]
\centering
\includegraphics[width=\textwidth]{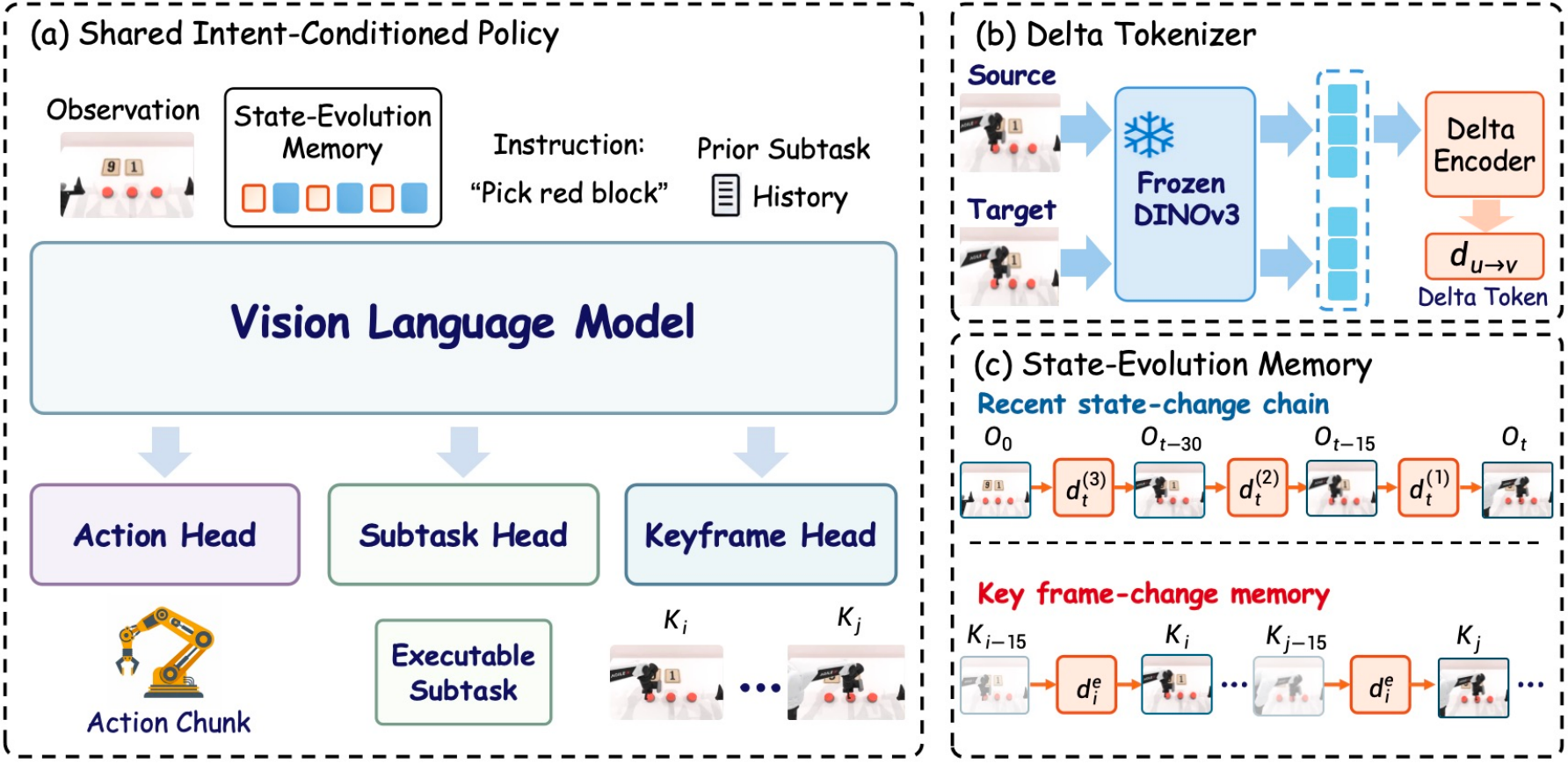}
\caption{\textbf{Architecture of EvoMem-VLA.} (a) A shared VLM performs action prediction, subtask generation, and keyframe selection from the current observation, state-evolution memory, instruction, and optional subtask history. Multi-stage tasks first generate an executable subtask as additional context for action generation. (b) The Delta Tokenizer maps an ordered frame pair to a directional, source-conditioned token. (c) State-Evolution Memory interleaves three recent delta-token groups with recent states and event-local delta tokens with selected keyframes.}
\label{fig:evomem_architecture}
\end{figure*}

EvoMem-VLA constructs state-evolution memory by organizing initial, recent, and selected keyframe observations alongside delta tokens that explicitly encode observed changes between frames. At each replanning step, a shared VLM backbone takes this memory, the current observation, and the task instruction as input and supports task-adaptive routing. For tasks without multi-stage planning, it predicts actions directly; for multi-stage tasks, it first generates an executable subtask and appends it to the input context to guide action generation. A lightweight Keyframe Head selects task-relevant observations to update the keyframe memory online. Figure~\ref{fig:evomem_architecture} summarizes the architecture.

\subsection{Preliminaries}
\label{sec:preliminaries}

\paragraph{Vision-Language-Action Models.}
At time $t$, let $\mathbf{o}_t$ be the robot observation, $\ell$ the task instruction, and $\mathbf{A}_t=[\mathbf{a}_t,\ldots,\mathbf{a}_{t+H-1}]\in\mathbb{R}^{H\times d_a}$ an action chunk with horizon $H$ and action dimension $d_a$. The visual encoder $E_v$ and language embedding layer $E_\ell$ produce $\mathbf{V}_t=E_v(\mathbf{o}_t)$ and $\mathbf{L}=E_\ell(\ell)$, forming $\mathbf{X}_t=[\mathbf{V}_t;\mathbf{L}]$. Given $H$ action-query tokens $\mathbf{Q}^{A}$, the VLM backbone $F_\theta$ and residual MLP action head $G_\psi$ predict
\begin{equation}
\mathbf{Z}_t^{A}=F_\theta([\mathbf{X}_t;\mathbf{Q}^{A}])_{\mathbf{Q}^{A}},
\qquad
\widehat{\mathbf{A}}_t=G_\psi(\mathbf{Z}_t^{A}).
\label{eq:vla_action}
\end{equation}
Here $[\,\cdot\,;\,\cdot\,]$ denotes token concatenation, and the subscript $\mathbf{Q}^{A}$ selects the $H$ action-query states. The parameters $\theta$ and $\psi$ belong to the backbone and action head, respectively. We optimize $\mathcal{L}_{\mathrm{act}}=\lVert\widehat{\mathbf{A}}_t-\mathbf{A}_t\rVert_1/(H d_a)$, where $\lVert\cdot\rVert_1$ sums element-wise absolute errors, and use $\Pi_{\theta,\psi}(\mathbf{X})$ to denote the complete action path in \Eqref{eq:vla_action}. EvoMem-VLA replaces the reactive visual context with the structured history defined below.

\paragraph{Conditional Delta Tokenization.}
Given an ordered source--target image pair $(\mathbf{I}_u,\mathbf{I}_v)$ with $u\leq v$, the frozen DINOv3 encoder $\phi$~\citep{dinov3} produces patch features $\mathbf{f}_u=\phi(\mathbf{I}_u)$ and $\mathbf{f}_v=\phi(\mathbf{I}_v)$. Following conditional feature compression in video~\citep{deltatok}, a paired-frame encoder $g_\Delta$ compresses the target information not explained by the source into a single token:
\begin{equation}
\mathbf{z}_{u\rightarrow v}=g_\Delta(\mathbf{f}_u,\mathbf{f}_v;\mathbf{q}_\Delta),
\label{eq:delta_tokenization}
\end{equation}
where $\mathbf{q}_\Delta$ is a learned query. A decoder reconstructs $\widehat{\mathbf{f}}_v=h_\Delta(\mathbf{f}_u,\mathbf{z}_{u\rightarrow v})$, and the tokenizer is pretrained with $\mathcal{L}_\Delta$, the log-cosh reconstruction loss averaged over patches and feature dimensions. This bottleneck yields a directional, source-conditioned change token rather than a pixel difference, optical-flow field, or action code. After pretraining, the DINOv3 encoder $\phi$ and $g_\Delta$ are frozen, $h_\Delta$ is discarded, and only the delta tokens are used by the VLA.

\subsection{State-Evolution Memory}
\label{sec:state_evolution_memory}

State-Evolution Memory links retained visual states to observed changes at complementary temporal scales: visual tokens preserve what was visible, while delta tokens describe how specific ordered frame pairs changed.

\paragraph{Multi-scale state evidence.}
At time $t$, fixed anchors contain observations at $0$, $t-30$, $t-15$, and $t$. The initial anchor preserves information that may disappear, while recent anchors represent the latest execution context. An ordered event-state bank $\{(\kappa_i,\mathbf{o}_{\kappa_i})\}_{i=1}^{m_t}$ retains $m_t\leq K$ selected observations with $\kappa_1<\cdots<\kappa_{m_t}\leq t$, where $K$ is its capacity. All historical indices are clamped to zero at episode start, reusing the initial frame when earlier observations are unavailable.

\paragraph{State-linked change tokens.}
The frozen Delta Tokenizer pairs each recent anchor with the observation 15 steps earlier, producing three token groups $\mathbf{d}_t^{(j)}$ ($j=1,2,3$, newest to oldest), with one token per observation stream. For each event state at $\kappa_i$, the tokenizer also encodes $(\kappa_i-\delta_e,\kappa_i)$, where $\delta_e$ is the event-local lag. A learned linear layer projects all delta tokens into the VLM embedding space; the same notation denotes the projected recent groups, and $\mathbf{d}_i^e$ denotes the projected primary-view event token. Source frames used only for tokenization are not added as RGB memory entries.

Let $\mathbf{V}_\tau$ be the visual tokens retained at time $\tau$ and
\[
\mathcal{M}_t^{\mathrm{anc}}=[\mathbf{V}_0;\mathbf{d}_t^{(3)};\mathbf{V}_{t-30};\mathbf{d}_t^{(2)};\mathbf{V}_{t-15};\mathbf{d}_t^{(1)};\mathbf{V}_t].
\]
Each delta token immediately precedes its target state, and event entries follow in capture order:
\begin{equation}
\mathcal{M}_t=\left[\mathcal{M}_t^{\mathrm{anc}};\bigl(\mathbf{d}_i^{e};\mathbf{V}_{\kappa_i}\bigr)_{i=1}^{m_t}\right].
\label{eq:state_evolution_memory}
\end{equation}
Every token represents its actual ordered pair; the model neither infers transitions between distant keyframes nor reconstructs unobserved intermediate states. Recent pairs are formed within each observation stream, whereas event states and event-local changes use the primary view.

The recent chain supplies regularly refreshed change evidence for ongoing execution. When keyframe memory is enabled, stored state--change pairs retain earlier evidence after it leaves the recent window. Pairing delta tokens with visual states keeps these changes tied to object and scene context, providing complementary information about what was observed and how it changed.

\paragraph{Online event-state acquisition.}
From the action-query states, the Keyframe Head predicts one recording probability per offset in the action horizon:
\begin{equation}
\widehat{\mathbf{p}}_t=\sigma\!\left(k_\omega(\mathbf{Z}_t^{A})\right)\in[0,1]^H.
\label{eq:keyframe_prediction}
\end{equation}
Here $k_\omega$ is the Keyframe Head with parameters $\omega$, and $\sigma$ is the sigmoid function. The output is indexed by offsets $j=0,\ldots,H-1$, corresponding to time $t+j$. Following EventVLA~\citep{eventvla}, the system schedules the highest-scoring eligible future offset only when its probability exceeds a recording threshold. The head predicts when to record, not the future observation itself. A scheduled observation is committed only when that time is reached. The bank retains the latest $K$ entries in chronological order.

\subsection{Task-Adaptive Routing}
\label{sec:task_adaptive_routing}

For tasks with temporally grounded subtask supervision~\citep{wla0}, EvoMem-VLA makes the current execution stage explicit. Let $s_t$ be the annotated subtask text for the action window, $R_t$ the prior subtask text, and $\mathbf{C}_t=[\mathcal{M}_t;\mathbf{L};E_\ell(R_t)]$. The subtask route is
\begin{equation}
\widehat{s}_t=\operatorname{Decode}_\theta(\mathbf{C}_t),
\qquad
\widehat{\mathbf{A}}_t=\Pi_{\theta,\psi}([\mathbf{C}_t;E_\ell(\widehat{s}_t)]).
\label{eq:subtask_route}
\end{equation}

\paragraph{Subtask-conditioned route.}
$\operatorname{Decode}_\theta$ uses the shared backbone and language-modeling head to generate the executable subtask. The generated text is appended to the same state--change context for a second backbone pass and action prediction. During training, $R_t$ comes from preceding subtask annotations; during rollout, it accumulates generated subtask text, skipping empty outputs and consecutive duplicates, and is reset at each episode. It records prior predictions, not verified task completion. Temporally aligned labels can repeat across calls or include consecutive descriptions within a window; the language end token only terminates text decoding.

\paragraph{Direct route.}
Task-adaptive routing is specified by task-level configuration rather than a learned complexity classifier. On the direct route, $R_t$ is empty and $\widehat{\mathbf{A}}_t=\Pi_{\theta,\psi}(\mathbf{C}_t)$. Both routes share the visual encoder, delta projection, VLM backbone, action queries, action head, and Keyframe Head; semantic supervision changes only the conditional path.

\subsection{Learning and Closed-Loop Inference}
\label{sec:learning_and_inference}

\paragraph{Two-stage optimization.}
Training separates change representation learning from policy learning. Stage 1 trains $g_\Delta$ and $h_\Delta$ on ordered frame pairs with $\phi$ frozen. Stage 2 freezes $\phi$ and $g_\Delta$ and optimizes the VLA, action head, delta projection, and Keyframe Head on robot demonstrations using
\begin{equation}
\mathcal{L}=\mathcal{L}_{\mathrm{act}}+\lambda_{\mathrm{lang}}\mathcal{L}_{\mathrm{lang}}+\lambda_{\mathrm{kf}}\mathcal{L}_{\mathrm{kf}}.
\label{eq:policy_objective}
\end{equation}
The weights $\lambda_{\mathrm{lang}}$ and $\lambda_{\mathrm{kf}}$ balance language and keyframe supervision. $\mathcal{L}_{\mathrm{act}}$ is applied to every sample. $\mathcal{L}_{\mathrm{lang}}$ is token-level cross-entropy over target subtask positions, with visual inputs, prompts, and action queries masked; direct-route samples contribute zero language loss. $\mathcal{L}_{\mathrm{kf}}$ is positive-weighted binary cross-entropy over supervised offsets and is zero without valid keyframe supervision. Each term is normalized over its valid elements before weighting. Subtask-route action training uses ground-truth subtask text, whereas deployment uses generated text. Because the tokenizer is frozen in Stage 2, $\mathcal{L}_\Delta$ is excluded from \Eqref{eq:policy_objective}.

\paragraph{Closed-loop inference.}
At each replanning step, the policy reads the fixed anchors, current observation, and committed event states; encodes the three recent delta groups and event-local delta tokens; and assembles $\mathcal{M}_t$ as in \Eqref{eq:state_evolution_memory}. The configured route then predicts an action chunk directly or generates and feeds back a subtask before action prediction. The Keyframe Head may concurrently schedule a future capture, which is written only when its timestamp is reached. The recent cache, bounded event-state bank, and enabled subtask history are then updated. This read--infer--act--write loop conditions every prediction only on evidence already observed by the robot.

\section{Experiments}
\label{sec:experiments}

\subsection{Experimental Setup}
\label{sec:experimental_setup}

\paragraph{Simulation setup.}
We evaluate EvoMem-VLA on two memory-dependent simulation benchmarks. RMBench~\citep{rmbench} comprises nine bimanual manipulation tasks, including five $M(1)$ tasks and four $M(n)$ tasks. Here, $M(1)$ tasks require retaining a single task-relevant past observation, whereas $M(n)$ tasks require information from multiple past observations. RoboMME~\citep{robomme} contains 16 tasks that evaluate counting, permanence, reference, and imitation memory. We implement EvoMem-VLA in the QwenOFT framework~\citep{starvla} and initialize the shared backbone from \texttt{Qwen3-VL-4B-Instruct}. For each benchmark, we jointly train a single policy across all tasks and evaluate 100 episodes per task; RoboMME uses 50 easy, 25 medium, and 25 hard episodes per task.

\paragraph{Real-world setup.}
We evaluate four tasks using a Franka single-arm robot and an AgileX COBOT dual-arm platform. We initialize the shared backbone from the robot-pretrained VLAct checkpoint~\citep{vlact} (\texttt{StarVLA/VLAct\_Qwen3\_Pretrain}) and train a separate policy for each embodiment. Each robot is evaluated on 10 episodes per task: 40 per embodiment and 80 in total. We report per-task success rates pooled across embodiments and their unweighted average. Camera layouts are shown in Figure~\ref{fig:robot_camera_setup} (Appendix~\ref{app:real_world_details}).

\paragraph{Compared baselines.}
For simulation, we compare against memory-free policies, including Diffusion Policy (DP)~\citep{diffusionpolicy}, ACT~\citep{act}, $\pi_{0.5}$~\citep{pi05}, X-VLA~\citep{xvla}, and QwenOFT; and memory-augmented policies, including MemER~\citep{memer}, Mem-0~\citep{rmbench}, MemoryVLA~\citep{memoryvla}, and EventVLA~\citep{eventvla}. For real-world experiments, we compare against $\pi_{0.5}$ and $\pi_{\mathrm{MEM}}$~\citep{mem}.

\subsection{Simulation Results}
\label{sec:simulation_results}

\begin{table}[t]
\centering
\caption{\textbf{RMBench results.} Success rates (\%) on the five $M(1)$ and four $M(n)$ tasks, together with group-wise and overall averages. EvoMem-VLA uses one policy jointly trained across all nine tasks. \textbf{Bold}: best; \underline{underlined}: second best.}
\label{tab:rmbench_results}
\scriptsize
\setlength{\tabcolsep}{2.6pt}
\renewcommand{\arraystretch}{1.10}
\resizebox{\textwidth}{!}{%
\begin{tabular}{lrrrrrrrrrrrr}
\toprule
\multirow{2}{*}{Method} & \multicolumn{6}{c}{$M(1)$} & \multicolumn{5}{c}{$M(n)$} & \multirow{2}{*}{\shortstack{Overall\\Avg.}} \\
\cmidrule(lr){2-7}\cmidrule(lr){8-12}
& \shortstack{Observe \&\\Pick Up} & \shortstack{Rearrange\\Blocks} & \shortstack{Put Back\\Block} & \shortstack{Swap\\Blocks} & \shortstack{Swap\\T} & Avg. & \shortstack{Battery\\Try} & \shortstack{Blocks\\Ranking Try} & \shortstack{Cover\\Blocks} & \shortstack{Press\\Button} & Avg. & \\
\midrule
\rowcolor{black!10}\multicolumn{13}{l}{$\blacktriangledown$\ \emph{Memory-free policies}} \\
DP~\citep{diffusionpolicy}                & 1.0  & 0.0  & 0.0  & 11.0 & 20.0 & 6.4  & 10.0 & 10.0 & 0.0  & 0.0  & 5.0  & 5.8 \\
ACT~\citep{act}                           & 1.0  & 29.0 & 0.0  & 2.0  & 2.0  & 6.8  & 19.0 & 0.0  & 0.0  & 0.0  & 4.8  & 5.9 \\
$\pi_{0.5}$~\citep{pi05}                  & 9.0  & 13.0 & 11.0 & 24.0 & 15.0 & 14.4 & 16.0 & 6.0  & 0.0  & 0.0  & 5.5  & 10.4 \\
X-VLA~\citep{xvla}                        & 9.0  & 13.0 & 18.0 & 16.0 & 3.0  & 11.8 & 26.0 & 1.0  & 2.0  & 0.0  & 7.3  & 9.8 \\
QwenOFT~\citep{starvla}                   & 0.0  & 0.0  & 0.0  & 0.0  & 0.0  & 0.0  & 14.0 & 37.0 & 0.0  & 0.0  & 12.8 & 5.7 \\
\midrule
\rowcolor{black!10}\multicolumn{13}{l}{$\blacktriangledown$\ \emph{Memory-augmented policies}} \\
MemER~\citep{memer}                       & 7.0  & 17.0 & 0.0  & 14.0 & 7.0  & 9.0  & 27.0 & 0.0  & 6.0  & 0.0  & 8.3  & 8.7 \\
Mem-0~\citep{rmbench}                     & 4.0  & 89.0 & 90.0 & 67.0 & 14.0 & 52.8 & 28.0 & 18.0 & 68.0 & 0.0  & 28.5 & 42.0 \\
MemoryVLA (OpenVLA)~\citep{memoryvla}     & 0.0  & 22.0 & 50.0 & 17.0 & 9.0  & 19.6 & 25.0 & 12.0 & 40.0 & 0.0  & 19.3 & 19.4 \\
MemoryVLA (QwenOFT)~\citep{memoryvla}     & 2.0  & 53.0 & 81.0 & 76.0 & 9.0  & 44.2 & 33.0 & 53.0 & 69.0 & 0.0  & 38.8 & 41.7 \\
EventVLA (VA only)~\citep{eventvla}       & \underline{21.0} & \underline{96.0} & \textbf{95.0} & \underline{96.0} & \underline{87.0} & \underline{79.0} & \underline{35.0} & \underline{81.0} & \textbf{97.0} & \underline{3.0} & \underline{54.0} & \underline{67.8} \\
\midrule
\textbf{EvoMem-VLA (Ours)}                & \textbf{28.0} & \textbf{100.0} & \underline{92.0} & \textbf{100.0} & \textbf{100.0} & \textbf{84.0} & \textbf{44.0} & \textbf{96.0} & \underline{96.0} & \textbf{70.0} & \textbf{76.5} & \textbf{80.7} \\
\bottomrule
\end{tabular}%
}
\end{table}

\paragraph{Results on RMBench.}
Table~\ref{tab:rmbench_results} reports task-level and aggregate success rates on RMBench. A single EvoMem-VLA policy jointly trained across all nine tasks achieves an overall success rate of 80.7\%, exceeding the previous best full-benchmark result of 67.8\% from EventVLA by 12.9 percentage points. EvoMem-VLA obtains 84.0\% on the $M(1)$ tasks and 76.5\% on the more demanding $M(n)$ subset. On $M(n)$, the jointly trained policy exceeds EventVLA's 54.0\% by 22.5 percentage points. At the task level, EvoMem-VLA ranks first or second on every task, reaching 100.0\% on \emph{Rearrange Blocks}, \emph{Swap Blocks}, and \emph{Swap T}, and 96.0\% on both \emph{Blocks Ranking Try} and \emph{Cover Blocks}. The largest gain over EventVLA occurs on \emph{Press Button}, where the success rate rises from 3.0\% to 70.0\%. The stronger improvement on $M(n)$ is consistent with the intended role of state-evolution memory and subtask conditioning in tracking progress through repeated and multi-stage interactions, while the $M(1)$ results show that the same jointly trained policy remains effective on tasks with simpler memory structure.

\begin{table}[t]
\centering
\caption{\textbf{RoboMME results.} Success rates (\%) across four memory categories and the overall average. EvoMem-VLA uses one policy jointly trained across all 16 tasks. For baselines with multiple variants, we report the highest-performing non-oracle variant. \textbf{Bold}: best; \underline{underlined}: second best.}
\label{tab:robomme_results}
\small
\setlength{\tabcolsep}{7pt}
\renewcommand{\arraystretch}{1.10}
\resizebox{\textwidth}{!}{%
\begin{tabular}{lrrrrr}
\toprule
Method & Counting & Permanence & Reference & Imitation & Avg. \\
\midrule
\rowcolor{black!10}\multicolumn{6}{l}{$\blacktriangledown$\ \emph{Memory-free policies}} \\
$\pi_{0.5}$~\citep{pi05}                       & 28.8 & 17.0 & 17.2 & 8.8  & 17.9 \\
\midrule
\rowcolor{black!10}\multicolumn{6}{l}{$\blacktriangledown$\ \emph{Memory-augmented policies}} \\
SimpleSG (QwenVL)~\citep{robomme}              & 44.6 & 19.6 & 25.2 & 26.6 & 29.0 \\
GroundSG (QwenVL)~\citep{robomme}              & 38.0 & 39.3 & 31.6 & 21.9 & 32.7 \\
TokenDrop (Modul)~\citep{robomme}              & 52.3 & 26.8 & 34.7 & 38.3 & 38.0 \\
FrameSamp (Modul)~\citep{robomme}              & \underline{65.2} & 25.1 & 36.3 & \underline{51.4} & \underline{44.5} \\
TTT (Expert)~\citep{robomme}                   & 36.0 & 22.9 & 19.7 & 10.8 & 22.3 \\
RMT (Modul)~\citep{robomme}                    & 34.1 & 15.8 & 21.1 & 9.7  & 20.2 \\
$\pi_{0.5}$ w/ past actions~\citep{robomme}    & 29.1 & 22.8 & 15.9 & 11.2 & 19.7 \\
MemER~\citep{memer}                           & 48.8 & \underline{53.2} & \underline{38.0} & 29.5 & 42.4 \\
\midrule
\textbf{EvoMem-VLA (Ours)}                    & \textbf{86.5} & \textbf{88.5} & \textbf{74.5} & \textbf{78.5} & \textbf{82.0} \\
\bottomrule
\end{tabular}%
}
\end{table}

\paragraph{Results on RoboMME.}
Table~\ref{tab:robomme_results} compares success rates across the four memory categories of RoboMME. A single EvoMem-VLA policy jointly trained across all 16 tasks achieves an overall success rate of 82.0\%, outperforming FrameSamp (Modul), the strongest baseline in the table at 44.5\%, by 37.5 percentage points, and MemER at 42.4\% by 39.6 points. EvoMem-VLA leads the compared baselines in every category, achieving 86.5\% on counting, 88.5\% on permanence, 74.5\% on reference, and 78.5\% on imitation. These results show consistent gains across all four evaluated memory capabilities.

\subsection{Real-World Experiments}
\label{sec:real_world_experiments}

\begin{figure}[!htbp]
\centering
\includegraphics[width=\textwidth]{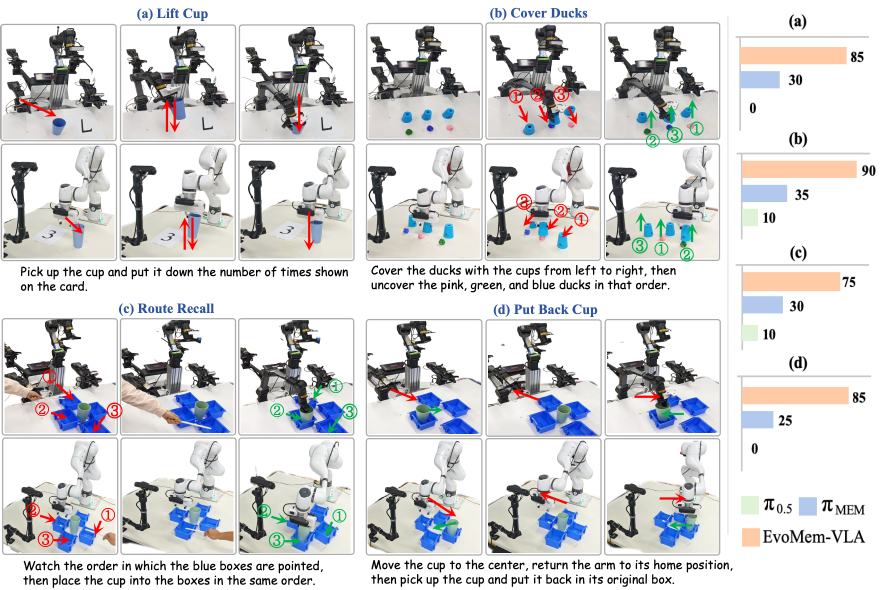}
\caption{\textbf{Real-world tasks and success rates.} Four tasks across two robot embodiments. Results on the right report each task's success rate (\%) averaged across the two embodiments.}
\label{fig:real_world_tasks_results}
\end{figure}

As shown in Figure~\ref{fig:real_world_tasks_results}, EvoMem-VLA achieves 83.8\% overall success, compared with 30.0\% for $\pi_{\mathrm{MEM}}$ and 5.0\% for $\pi_{0.5}$, and leads both baselines on all four tasks. \emph{Lift Cup} tests repetition counting (85.0\%), \emph{Cover Ducks} tests object-identity memory under occlusion (90.0\%), \emph{Route Recall} tests memory for demonstrated spatial order (75.0\%), and \emph{Put Back Cup} tests recall of an object's original location after displacement (85.0\%).

These tasks require different uses of interaction history. In \emph{Lift Cup}, the cup returns to a similar pose after each repetition, so its current position does not indicate how many lifts remain. \emph{Cover Ducks} requires remembering duck identities after they are covered, whereas \emph{Route Recall} requires retaining the order of a transient pointing demonstration, including repeated visits, and tracking which placements are complete. In \emph{Put Back Cup}, the shared intermediate position no longer identifies the original box. Together with the RMBench ablations, these results support conditional delta tokenization as an effective way to preserve interaction-progress evidence alongside historical states for robot control.

\begin{figure*}[!t]
\centering
\includegraphics[width=\textwidth]{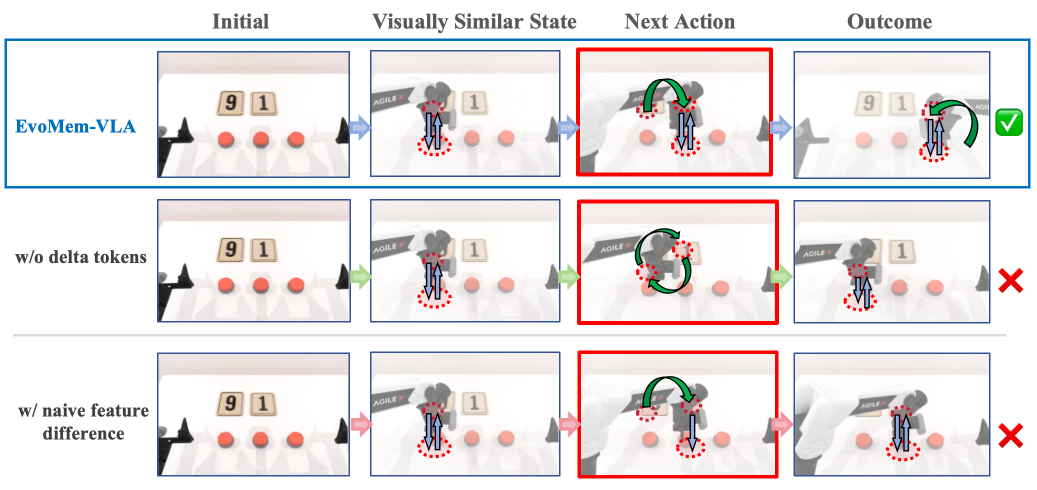}
\caption{\textbf{Trajectory comparison on RMBench \emph{Press Button}.} Under the same episode initialization, EvoMem-VLA completes the task; the variant without delta tokens follows an incorrect trajectory, while the naive feature difference variant repeats actions. Red boxes highlight the next actions.}
\label{fig:rmbench_change_trajectory_comparison}
\end{figure*}

\subsection{Ablation Study}
\label{sec:ablation_study}

\begin{table}[t]
\centering
\caption{\textbf{Component ablations on RMBench.} Success rates (\%) averaged over the five $M(1)$ tasks, the four $M(n)$ tasks, and all nine tasks. \textbf{Bold}: best; \underline{underlined}: second best. All variants are implemented and evaluated within our framework under the same training and evaluation protocol.}
\label{tab:rmbench_ablation}
\small
\setlength{\tabcolsep}{12pt}
\renewcommand{\arraystretch}{1.12}
\begin{tabular}{lccc}
\toprule
Method & $M(1)$ & $M(n)$ & Overall \\
\midrule
\textbf{EvoMem-VLA}                 & \textbf{84.0}   & \textbf{76.5} & \textbf{80.7} \\
w/ naive feature difference         & 52.0            & 41.0           & 47.1 \\
w/o subtask route                   & \textbf{84.0}   & 33.0           & \underline{61.3} \\
w/o delta tokens                    & \underline{59.0} & \underline{44.0} & 52.3 \\
w/o delta tokens \& subtask route   & 55.0            & 31.0           & 44.3 \\
\bottomrule
\end{tabular}
\end{table}

\paragraph{Effect of conditional delta tokenization.}
Table~\ref{tab:rmbench_ablation} shows that removing delta tokens reduces overall success from 80.7\% to 52.3\%. This 28.4-point drop exceeds the 19.4-point drop from removing subtask routing, identifying conditional delta tokenization as the larger contributor to overall performance in these ablations. Removing delta tokens lowers success on both $M(1)$ (84.0\% to 59.0\%) and $M(n)$ (76.5\% to 44.0\%). The naive feature difference variant replaces the learned Delta Encoder with target-minus-source feature subtraction followed by projection into the VLM embedding space. It achieves only 47.1\%, below even the no-delta variant. The full model's 33.6-point advantage over naive differences shows that exposing feature differences alone is insufficient: learning source-conditioned changes is central to constructing useful memory for action generation.

\paragraph{Complementary role of subtask conditioning.}
As shown in Table~\ref{tab:rmbench_ablation}, removing subtask routing leaves $M(1)$ success unchanged at 84.0\% but reduces $M(n)$ success from 76.5\% to 33.0\%, highlighting its role in multi-stage execution. Removing both components yields 44.3\% overall success. From this state-only variant, restoring delta tokens alone raises success by 17.0 points to 61.3\%, whereas restoring routing alone yields an 8.0-point gain to 52.3\%. With both components, success reaches 80.7\%. Thus, delta tokens provide the larger overall gain with or without routing, while subtask conditioning complements this change evidence for multi-stage control.

\paragraph{Qualitative trajectory comparison.}
In the \emph{Press Button} episode in Figure~\ref{fig:rmbench_change_trajectory_comparison}, the instruction requires pressing the left button nine times, the middle button once, and finally the rightmost termination button. Repeated presses leave visually similar states, so the policy must remember the completed press count to decide when to switch buttons. Under the same initialization, EvoMem-VLA completes this sequence, whereas the variant without delta tokens follows an incorrect trajectory and returns to the left button. The naive feature difference variant reaches the middle button but presses it repeatedly instead of proceeding to termination. These trajectories illustrate how conditional delta tokens preserve the interaction-progress evidence needed to select the appropriate next action.

\section{Conclusion}

In this paper, we presented EvoMem-VLA, a state-evolution memory paradigm that explicitly retains observed interaction changes alongside historical states for action generation. Our unified VLA uses conditional delta tokenization to construct multi-scale state-evolution memory and supports direct or subtask-conditioned action generation through task-adaptive routing. Experiments on RMBench, RoboMME, and four real-world tasks across two robot embodiments demonstrate improvements over the previous state of the art. RMBench ablations further validate the central contribution of conditional delta tokenization.

\subsection*{AI use statement}

Generative AI tools were used for literature search, translation, and language refinement, and to provide feedback on the research methodology and experimental evaluation. The authors reviewed all AI-assisted content and verified the references and associated claims. The authors take full responsibility for the final manuscript, including its scientific claims, analyses, and editorial decisions.

\subsection*{Reproducibility statement}

Implementation and training details, memory construction, evaluation protocols, baseline sources, and controlled ablation settings are documented in Appendices~\ref{app:implementation}--\ref{app:ablation_details}. The anonymized supplementary material provides the principal training and evaluation code together with real-world demonstration videos.

\bibliography{references}

@misc{rt2,
  author = {Brohan, Anthony and others},
  title = {{RT-2: Vision-Language-Action Models Transfer Web Knowledge to Robotic Control}},
  year = {2023},
  eprint = {2307.15818},
  archivePrefix = {arXiv},
  url = {https://arxiv.org/abs/2307.15818}
}

@misc{openvla,
  author = {Kim, Moo Jin and others},
  title = {{OpenVLA: An Open-Source Vision-Language-Action Model}},
  year = {2024},
  eprint = {2406.09246},
  archivePrefix = {arXiv},
  url = {https://arxiv.org/abs/2406.09246}
}

@misc{pi0,
  author = {Black, Kevin and others},
  title = {{$\pi_{0}$: A Vision-Language-Action Flow Model for General Robot Control}},
  year = {2024},
  eprint = {2410.24164},
  archivePrefix = {arXiv},
  url = {https://arxiv.org/abs/2410.24164}
}

@misc{xvla,
  author = {Zheng, Jinliang and others},
  title = {{X-VLA: Soft-Prompted Transformer as Scalable Cross-Embodiment Vision-Language-Action Model}},
  year = {2025},
  eprint = {2510.10274},
  archivePrefix = {arXiv},
  url = {https://arxiv.org/abs/2510.10274}
}

@misc{rmbench,
  author = {Chen, Tianxing and others},
  title = {{RMBench: Memory-Dependent Robotic Manipulation Benchmark with Insights into Policy Design}},
  year = {2026},
  eprint = {2603.01229},
  archivePrefix = {arXiv},
  url = {https://arxiv.org/abs/2603.01229}
}

@inproceedings{hamlet,
  author = {Koo, Myungkyu and others},
  title = {{HAMLET: Switch your Vision-Language-Action Model into a History-Aware Policy}},
  booktitle = {ICLR},
  year = {2026},
  eprint = {2510.00695},
  archivePrefix = {arXiv},
  url = {https://arxiv.org/abs/2510.00695}
}

@inproceedings{memoryvla,
  author = {Shi, Hao and others},
  title = {{MemoryVLA: Perceptual-Cognitive Memory in Vision-Language-Action Models for Robotic Manipulation}},
  booktitle = {ICLR},
  year = {2026},
  eprint = {2508.19236},
  archivePrefix = {arXiv},
  url = {https://arxiv.org/abs/2508.19236}
}

@misc{eventvla,
  author = {Yang, Ganlin and others},
  title = {{EventVLA: Event-Driven Visual Evidence Memory for Long-Horizon Vision-Language-Action Policies}},
  year = {2026},
  eprint = {2606.20092},
  archivePrefix = {arXiv},
  url = {https://arxiv.org/abs/2606.20092}
}

@misc{pi05,
  author = {{Physical Intelligence} and others},
  title = {{$\pi_{0.5}$: a Vision-Language-Action Model with Open-World Generalization}},
  year = {2025},
  eprint = {2504.16054},
  archivePrefix = {arXiv},
  url = {https://arxiv.org/abs/2504.16054}
}

@article{diffusionpolicy,
  author = {Chi, Cheng and Xu, Zhenjia and Feng, Siyuan and Cousineau, Eric and Du, Yilun and Burchfiel, Benjamin and Tedrake, Russ and Song, Shuran},
  title = {{Diffusion Policy: Visuomotor Policy Learning via Action Diffusion}},
  journal = {The International Journal of Robotics Research},
  volume = {44},
  number = {10--11},
  pages = {1684--1704},
  year = {2025},
  eprint = {2303.04137},
  archivePrefix = {arXiv},
  url = {https://arxiv.org/abs/2303.04137}
}

@misc{act,
  author = {Zhao, Tony Z. and Kumar, Vikash and Levine, Sergey and Finn, Chelsea},
  title = {{Learning Fine-Grained Bimanual Manipulation with Low-Cost Hardware}},
  year = {2023},
  eprint = {2304.13705},
  archivePrefix = {arXiv},
  url = {https://arxiv.org/abs/2304.13705}
}

@misc{vlact,
  author = {Yang, Senqiao and Wang, Chengyao and Chen, Yuxin and Wang, Zixuan and Tang, Longxiang and Gui, Haokun and Ye, Jinhui and Lu, Changsheng and Wu, Xiaoyang and Zhu, Mingkang and Chen, Pengguang and Liu, Shu and Tian, Zhuotao and Zhao, Hengshuang and Yu, Bei and Jia, Jiaya},
  title = {{Beyond Data Scaling: Representation-Centric Continued Pre-training for Vision-Language-Action Models}},
  year = {2026},
  eprint = {2608.27550},
  archivePrefix = {arXiv},
  url = {https://arxiv.org/abs/2608.27550}
}

@misc{mem,
  author = {Torne, Marcel and Pertsch, Karl and Walke, Homer and Vedder, Kyle and Nair, Suraj and Ichter, Brian and Ren, Allen Z. and Wang, Haohuan and Tang, Jiaming and Stachowicz, Kyle and Dhabalia, Karan and Equi, Michael and Vuong, Quan and Springenberg, Jost Tobias and Levine, Sergey and Finn, Chelsea and Driess, Danny},
  title = {{MEM: Multi-Scale Embodied Memory for Vision Language Action Models}},
  year = {2026},
  eprint = {2603.03596},
  archivePrefix = {arXiv},
  url = {https://arxiv.org/abs/2603.03596}
}

@misc{starvla,
  author = {{StarVLA Community}},
  title = {{StarVLA: A Lego-like Codebase for Vision-Language-Action Model Developing}},
  year = {2026},
  eprint = {2604.05014},
  archivePrefix = {arXiv},
  url = {https://arxiv.org/abs/2604.05014}
}

@misc{robomme,
  author = {Dai, Yinpei and Fu, Hongze and Lee, Jayjun and Liu, Yuejiang and Zhang, Haoran and Yang, Jianing and Finn, Chelsea and Fazeli, Nima and Chai, Joyce},
  title = {{RoboMME: Benchmarking and Understanding Memory for Robotic Generalist Policies}},
  year = {2026},
  eprint = {2603.04639},
  archivePrefix = {arXiv},
  url = {https://arxiv.org/abs/2603.04639}
}

@misc{wla0,
  author = {Yang, Yi and others},
  title = {{World-Language-Action Model for Unified World Modeling, Language Reasoning, and Action Synthesis}},
  year = {2026},
  eprint = {2606.05979},
  archivePrefix = {arXiv},
  url = {https://arxiv.org/abs/2606.05979}
}

@misc{roboflamingo,
  author = {Li, Xinghang and others},
  title = {{Vision-Language Foundation Models as Effective Robot Imitators}},
  year = {2023},
  eprint = {2311.01378},
  archivePrefix = {arXiv},
  url = {https://arxiv.org/abs/2311.01378}
}

@inproceedings{tracevla,
  author = {Zheng, Ruijie and others},
  title = {{TraceVLA: Visual Trace Prompting Enhances Spatial-Temporal Awareness for Generalist Robotic Policies}},
  booktitle = {ICLR},
  year = {2025},
  eprint = {2412.10345},
  archivePrefix = {arXiv},
  url = {https://arxiv.org/abs/2412.10345}
}

@misc{fourdvla,
  author = {Zhang, Jiahui and others},
  title = {{4D-VLA: Spatiotemporal Vision-Language-Action Pretraining with Cross-Scene Calibration}},
  year = {2025},
  eprint = {2506.22242},
  archivePrefix = {arXiv},
  url = {https://arxiv.org/abs/2506.22242}
}

@misc{ecot,
  author = {Zawalski, Micha{\l} and others},
  title = {{Robotic Control via Embodied Chain-of-Thought Reasoning}},
  year = {2024},
  eprint = {2407.08693},
  archivePrefix = {arXiv},
  url = {https://arxiv.org/abs/2407.08693}
}

@inproceedings{memer,
  author = {Sridhar, Ajay and Pan, Jennifer and Sharma, Satvik and Finn, Chelsea},
  title = {{Scaling up Memory for Robotic Control via Experience Retrieval}},
  booktitle = {ICLR},
  year = {2026},
  url = {https://proceedings.iclr.cc/paper_files/paper/2026/hash/9da515b1ad19d032a7398f00f5ff9b0c-Abstract-Conference.html}
}

@misc{cogact,
  author = {Li, Qixiu and Liang, Yaobo and Wang, Zeyu and Luo, Lin and Chen, Xi and Liao, Mozheng and Wei, Fangyun and Deng, Yu and Xu, Sicheng and Zhang, Yizhong and others},
  title = {{CogACT: A Foundational Vision-Language-Action Model for Synergizing Cognition and Action in Robotic Manipulation}},
  year = {2024},
  eprint = {2411.19650},
  archivePrefix = {arXiv},
  url = {https://arxiv.org/abs/2411.19650}
}

@misc{dexvla,
  author = {Wen, Junjie and Zhu, Yichen and Li, Jinming and Tang, Zhibin and Shen, Chaomin and Feng, Feifei},
  title = {{DexVLA: Vision-Language Model with Plug-In Diffusion Expert for General Robot Control}},
  year = {2025},
  eprint = {2502.05855},
  archivePrefix = {arXiv},
  url = {https://arxiv.org/abs/2502.05855}
}

@misc{hybridvla,
  author = {Liu, Jiaming and Chen, Hao and An, Pengju and Liu, Zhuoyang and Zhang, Renrui and Gu, Chenyang and Li, Xiaoqi and Guo, Ziyu and Chen, Sixiang and Liu, Mengzhen and others},
  title = {{HybridVLA: Collaborative Diffusion and Autoregression in a Unified Vision-Language-Action Model}},
  year = {2025},
  eprint = {2503.10631},
  archivePrefix = {arXiv},
  url = {https://arxiv.org/abs/2503.10631}
}

@misc{kcvla,
  author = {Chen, Yipeng and others},
  title = {{Non-Markovian Long-Horizon Robot Manipulation via Keyframe Chaining}},
  year = {2026},
  eprint = {2603.01465},
  archivePrefix = {arXiv},
  url = {https://arxiv.org/abs/2603.01465}
}

@misc{swiftvla,
  author = {Ni, Chaojun and others},
  title = {{SwiftVLA: Unlocking Spatiotemporal Dynamics for Lightweight VLA Models at Minimal Overhead}},
  year = {2025},
  eprint = {2512.00903},
  archivePrefix = {arXiv},
  url = {https://arxiv.org/abs/2512.00903}
}

@inproceedings{thinkact,
  author = {Huang, Chi-Pin and others},
  title = {{ThinkAct: Vision-Language-Action Reasoning via Reinforced Visual Latent Planning}},
  booktitle = {NeurIPS},
  year = {2025},
  eprint = {2507.16815},
  archivePrefix = {arXiv},
  url = {https://arxiv.org/abs/2507.16815}
}

@inproceedings{deltatok,
  author = {Kerssies, Tommie and others},
  title = {{A Frame is Worth One Token: Efficient Generative World Modeling with Delta Tokens}},
  booktitle = {CVPR},
  year = {2026},
  url = {https://openaccess.thecvf.com/content/CVPR2026/papers/Kerssies_A_Frame_is_Worth_One_Token_Efficient_Generative_World_Modeling_CVPR_2026_paper.pdf}
}

@misc{dinov3,
  author = {Sim{\'e}oni, Oriane and Vo, Huy V. and Seitzer, Maximilian and Baldassarre, Federico and Oquab, Maxime and Jose, Cijo and Khalidov, Vasil and Szafraniec, Marc and Yi, Seungeun and Ramamonjisoa, Micha{\"e}l and others},
  title = {{DINOv3}},
  year = {2025},
  eprint = {2508.10104},
  archivePrefix = {arXiv},
  url = {https://arxiv.org/abs/2508.10104}
}

@inproceedings{openvlaoft,
  author = {Kim, Moo Jin and Finn, Chelsea and Liang, Percy},
  title = {{Fine-Tuning Vision-Language-Action Models: Optimizing Speed and Success}},
  booktitle = {Robotics: Science and Systems},
  year = {2025},
  eprint = {2502.19645},
  archivePrefix = {arXiv},
  url = {https://arxiv.org/abs/2502.19645}
}

@misc{spatialvla,
  author = {Qu, Delin and Song, Haoming and Chen, Qizhi and Yao, Yuanqi and Ye, Xinyi and Ding, Yan and Wang, Zhigang and Gu, JiaYuan and Zhao, Bin and Wang, Dong and Li, Xuelong},
  title = {{SpatialVLA: Exploring Spatial Representations for Visual-Language-Action Model}},
  year = {2025},
  eprint = {2501.15830},
  archivePrefix = {arXiv},
  url = {https://arxiv.org/abs/2501.15830}
}

@misc{smolvla,
  author = {Shukor, Mustafa and Aubakirova, Dana and Capuano, Francesco and Kooijmans, Pepijn and Palma, Steven and Zouitine, Adil and Aractingi, Michel and Pascal, Caroline and Russi, Martino and Marafioti, Andres and Alibert, Simon and Cord, Matthieu and Wolf, Thomas and Cadene, Remi},
  title = {{SmolVLA: A Vision-Language-Action Model for Affordable and Efficient Robotics}},
  year = {2025},
  eprint = {2506.01844},
  archivePrefix = {arXiv},
  url = {https://arxiv.org/abs/2506.01844}
}

@misc{rth,
  author = {Belkhale, Suneel and Ding, Tianli and Xiao, Ted and Sermanet, Pierre and Vuong, Quon and Tompson, Jonathan and Chebotar, Yevgen and Dwibedi, Debidatta and Sadigh, Dorsa},
  title = {{RT-H: Action Hierarchies Using Language}},
  year = {2024},
  eprint = {2403.01823},
  archivePrefix = {arXiv},
  url = {https://arxiv.org/abs/2403.01823}
}

@misc{sam2act,
  author = {Fang, Haoquan and Grotz, Markus and Pumacay, Wilbert and Wang, Yi Ru and Fox, Dieter and Krishna, Ranjay and Duan, Jiafei},
  title = {{SAM2Act: Integrating Visual Foundation Model with A Memory Architecture for Robotic Manipulation}},
  year = {2025},
  eprint = {2501.18564},
  archivePrefix = {arXiv},
  url = {https://arxiv.org/abs/2501.18564}
}
\bibliographystyle{iclr2027_conference}

\clearpage
\appendix
\section{Implementation and Training Details}
\label{app:implementation}

\subsection{Policy Architecture and Initialization}
\label{app:policy_architecture}

We implement EvoMem-VLA using the QwenOFT framework in StarVLA~\citep{starvla}. Simulation policies are initialized from \texttt{Qwen3-VL-4B-Instruct}, with one policy jointly trained across all tasks in each benchmark. For real-world experiments, we initialize from the robot-pretrained \texttt{StarVLA/VLAct\_Qwen3\_Pretrain} checkpoint~\citep{vlact} and train one policy per embodiment. The continuous action head is a residual MLP with two residual blocks and a hidden width twice that of the VLM. A learned linear layer maps delta tokens into the VLM embedding space. Subtask generation uses the backbone's language-modeling head, and route selection is specified by the task configuration.

\paragraph{Simulation route assignments.}
RMBench uses the direct route for its five $M(1)$ tasks and the subtask-conditioned route for its four $M(n)$ tasks (Battery Try, Blocks Ranking Try, Cover Blocks, and Press Button). All 16 RoboMME tasks use the subtask-conditioned route. These assignments remain fixed during training and evaluation.

\paragraph{Supervision and alignment.}
RMBench uses episode-level subtask text segments aligned by frame index: preceding segments form the history, and segments overlapping the current action window supply the target text. RoboMME uses its native subgoal text and subgoal-boundary annotations for subtask and keyframe supervision, aligned with observations and actions by episode timestep. The current subgoal supplies the text target, while earlier subgoals form the history. These labels are used during training; deployment uses model-generated subtasks and keyframe predictions.

\subsection{Delta Tokenizer Training}
\label{app:delta_tokenizer}

The Delta Tokenizer follows DeltaTok~\citep{deltatok} and extracts patch features with frozen DINOv3~\citep{dinov3}. For simulation, we initialize from the released DeltaTok weights and adapt a separate tokenizer on each benchmark's demonstration dataset. Source and target images come from the same camera stream, separated by 15 observation steps. The encoder compresses each pair into one delta token, and the decoder reconstructs the target features conditioned on the source features and this token. Only the tokenizer encoder and decoder are updated, using the mean log-cosh reconstruction loss; DINOv3 remains frozen. Table~\ref{tab:tokenizer_training_details} summarizes the shared implementation and optimization settings.

Frame pairs are sampled within demonstration episodes, never across episode boundaries. RoboMME adaptation uses the same released demonstration episodes as policy training, with task-balanced sampling. Tokenizer adaptation is completed before policy training; the 80,000-step policy budget does not include tokenizer updates.

Real-world experiments use the official pretrained DeltaTok checkpoint directly, without task-specific tokenizer adaptation. During policy training and deployment, both DINOv3 and the Delta Tokenizer remain frozen, and the reconstruction decoder is discarded.

\begin{table}[!ht]
\centering
\caption{\textbf{Delta Tokenizer implementation and adaptation settings for the simulation benchmarks.}}
\label{tab:tokenizer_training_details}
\small
\begin{tabular}{ll}
\toprule
Setting & Value \\
\midrule
Feature extractor & DINOv3 ViT-B/16, frozen \\
Tokenizer input resolution & $256\times256$ \\
Encoder / decoder depth & 12 / 12 layers \\
Delta-token dimension & 768 \\
Delta tokens per frame pair & 1 \\
Source--target interval & 15 observation steps \\
GPUs & 8 \\
Batch size per GPU / global batch size & 16 / 128 \\
Learning rate & $10^{-4}$ \\
Weight decay & $10^{-2}$ \\
Gradient clipping threshold & $10^{-2}$ \\
Training precision & bfloat16 \\
\bottomrule
\end{tabular}
\end{table}

\subsection{Simulation Policy Training}
\label{app:simulation_training}

Table~\ref{tab:simulation_training_details} summarizes policy training on RMBench and RoboMME. Both benchmarks use an 80,000-step training budget with eight GPUs, a per-GPU batch size of four, and no gradient accumulation. We use AdamW with $\beta=(0.9,0.95)$ and $\epsilon=10^{-8}$. The learning rate follows a cosine schedule after 5,000 warmup steps, with a minimum learning rate of $5\times10^{-7}$. The backbone, delta projection, and action head are trained while the change representation remains frozen.

All samples supervise action prediction. Subtask-route samples additionally supervise the annotated subtask text with a language-loss weight of $0.005$; direct-route samples contribute no language loss. Action training uses the annotated subtask, whereas deployment uses the generated subtask. On RoboMME, the Keyframe Head is trained with positive-weighted binary cross-entropy, using a positive-class weight of 7 and a keyframe-loss weight of 1. These optimization settings apply to the simulation policies.

\begin{table}[!ht]
\centering
\caption{\textbf{Policy configuration and optimization settings for the simulation benchmarks.}}
\label{tab:simulation_training_details}
\small
\begin{tabular}{lll}
\toprule
Setting & RMBench & RoboMME \\
\midrule
Camera streams & Head and two wrists & Front and wrist \\
Policy image resolution & $224\times224$ & $224\times224$ \\
Action dimension & 14 & 8 \\
Action-chunk horizon & 50 & 20 \\
Recent delta-token groups & 3 & 3 \\
Keyframe memory & Disabled & Enabled \\
Keyframe capacity & Not used & 16 \\
Keyframe recording threshold & Not used & 0.55 \\
Training budget & 80,000 steps & 80,000 steps \\
GPUs & 8 & 8 \\
Batch size per GPU & 4 & 4 \\
Gradient accumulation steps & 1 & 1 \\
Global batch size & 32 & 32 \\
Backbone learning rate & $10^{-5}$ & $10^{-5}$ \\
Action-head learning rate & $10^{-4}$ & $10^{-4}$ \\
Optimizer weight decay & $10^{-8}$ & $10^{-8}$ \\
Gradient clipping threshold & 1.0 & 1.0 \\
Warmup steps & 5,000 & 5,000 \\
Random seed & 42 & 42 \\
\bottomrule
\end{tabular}
\end{table}

\subsection{Memory Construction and Episode Boundaries}
\label{app:memory_construction}

The three recent delta-token groups encode $(t-45,t-30)$, $(t-30,t-15)$, and $(t-15,t)$, respectively, with one token per camera stream. The initial frame is retained separately as a visual anchor. Negative indices are clamped to zero, reusing the initial frame at the beginning of an episode. Frames used only as tokenizer sources are not added as extra RGB memory entries.

RMBench uses the initial and recent observations together with these three delta-token groups; keyframe memory is disabled. RoboMME additionally retains up to 16 keyframes, each paired with a primary-view delta token computed from the observation 15 steps earlier. Its Keyframe Head schedules a recording when the predicted probability exceeds 0.55. The observation is stored only when its scheduled capture time is reached, and capacity overflow removes the oldest record. Visual memory and prior subtask history are reset between episodes.

\subsection{Real-World Training and Deployment}
\label{app:real_world_details}

\begin{figure}[!t]
\centering
\begin{minipage}[b]{0.49\textwidth}
\centering
\includegraphics[width=0.80\linewidth]{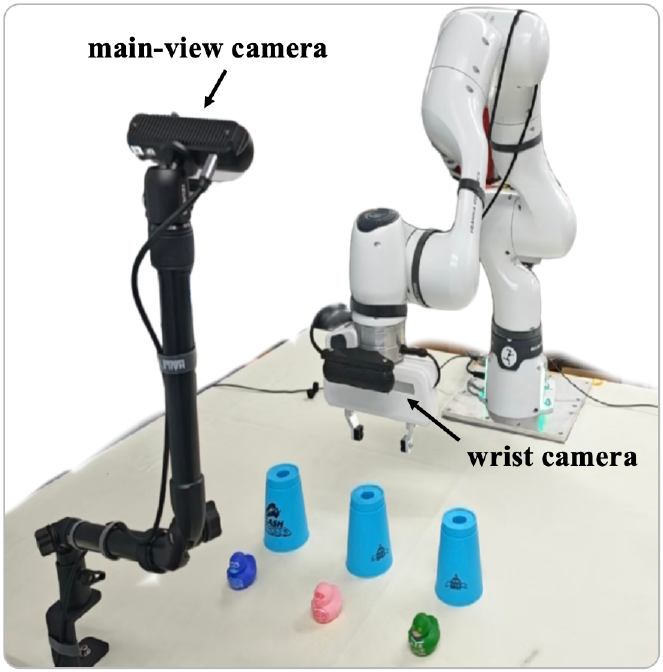}
\par\smallskip
{\small (a) Franka single-arm robot}
\end{minipage}\hfill
\begin{minipage}[b]{0.49\textwidth}
\centering
\includegraphics[width=0.80\linewidth]{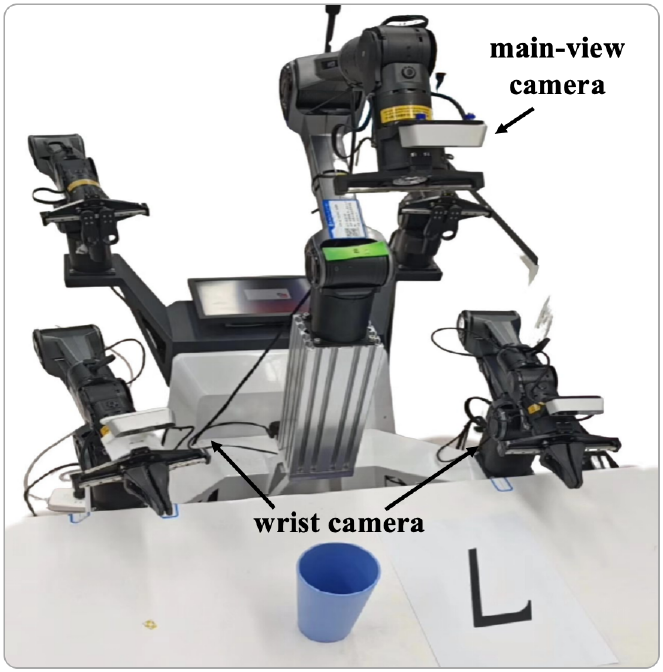}
\par\smallskip
{\small (b) AgileX COBOT dual-arm platform}
\end{minipage}
\caption{\textbf{Camera configurations of the two real-world robot platforms.} Main-view and wrist cameras are marked.}
\label{fig:robot_camera_setup}
\end{figure}

We use a Franka single-arm robot and an AgileX COBOT dual-arm platform. Figure~\ref{fig:robot_camera_setup} shows their camera configurations. The Franka setup provides one main-view camera and one wrist camera; the AgileX COBOT setup provides one main-view camera and two wrist cameras, one for each arm.

We collect 50 demonstrations per task for each embodiment, yielding 200 demonstrations per embodiment and 400 in total across the four tasks. Each embodiment has a separately trained policy initialized as described in Appendix~\ref{app:policy_architecture}. We fine-tune all VLA policy parameters while keeping DINOv3 and the official pretrained Delta Tokenizer frozen. Both embodiments use the same optimization settings, following the real-world training configuration of EventVLA~\citep{eventvla}. Specifically, we use AdamW for 60,000 optimization steps with a global batch size of 32 and a peak learning rate of $5\times10^{-5}$. After 2,000 warmup steps, the learning rate follows a cosine decay schedule to $5\times10^{-6}$. Table~\ref{tab:real_world_training_details} summarizes the training and deployment settings.

At each replanning step, the policy predicts an action chunk of 30 steps and executes its first 16 steps before replanning. Keyframe memory is enabled with a capacity of five frames, alongside the initial and recent observations and three recent delta-token groups.

\begin{table}[!t]
\centering
\caption{\textbf{Real-world training and deployment settings shared by both embodiments.} Optimization settings follow EventVLA; initialization, action horizons, and memory construction follow our implementation.}
\label{tab:real_world_training_details}
\small
\begin{tabular}{ll}
\toprule
Setting & Value \\
\midrule
Demonstrations per task per embodiment & 50 \\
Demonstrations per embodiment & 200 \\
Policy initialization & \texttt{StarVLA/VLAct\_Qwen3\_Pretrain} \\
Policy fine-tuning & Full-parameter \\
Delta Tokenizer & Official pretrained checkpoint, frozen \\
Optimizer & AdamW \\
AdamW betas / epsilon & $(0.9,0.95)$ / $10^{-8}$ \\
Weight decay & 0.01 \\
Optimization steps & 60,000 \\
Global batch size & 32 \\
Peak learning rate & $5\times10^{-5}$ \\
Warmup steps & 2,000 \\
Learning-rate schedule & Cosine decay \\
Minimum learning rate & $5\times10^{-6}$ \\
Training precision & bfloat16 \\
Camera inputs & Main-view and wrist cameras \\
Predicted action-chunk length & 30 steps \\
Executed actions before replanning & 16 steps \\
Recent delta-token groups & 3 \\
Keyframe memory capacity & 5 frames \\
\bottomrule
\end{tabular}
\end{table}

\section{Tasks and Evaluation Protocols}
\label{app:evaluation_protocols}

\subsection{Simulation Benchmarks}
\label{app:simulation_evaluation}

We use the official task definitions, environment settings, and success criteria of RMBench~\citep{rmbench} and RoboMME~\citep{robomme}, without modifying the benchmark environments. Each benchmark uses one jointly trained EvoMem-VLA policy. We evaluate 100 episodes per task, totaling 900 episodes on RMBench and 1,600 on RoboMME. Each RoboMME task includes 50 easy, 25 medium, and 25 hard episodes.

For RMBench, Table~\ref{tab:rmbench_results} reports each task's success rate and group-wise and overall aggregates. We compute these as unweighted means over the five $M(1)$ tasks, the four $M(n)$ tasks, and all nine tasks, respectively, except that EventVLA's overall score retains the published value of 67.8\%. The all-task mean weights the two groups by their task counts, rather than averaging the two group means equally. For RoboMME, Table~\ref{tab:robomme_results} averages task success rates within each memory category and across all 16 tasks. Reported success rates are rounded to one decimal place.

\subsection{Real-World Tasks and Success Criteria}
\label{app:real_world_evaluation}

Table~\ref{tab:real_world_task_protocols} specifies the four tasks in Figure~\ref{fig:real_world_tasks_results}, evaluated on both platforms with the camera layouts in Figure~\ref{fig:robot_camera_setup}. Human pointing in Route Recall supplies the task input; no task allows human corrections or assistance during robot execution.

\begin{table}[!ht]
\centering
\caption{\textbf{Real-world task instructions and success criteria.}}
\label{tab:real_world_task_protocols}
\small
\setlength{\tabcolsep}{4pt}
\renewcommand{\arraystretch}{1.15}
\begin{tabular}{@{}>{\raggedright\arraybackslash}p{0.15\textwidth}>{\raggedright\arraybackslash}p{0.38\textwidth}>{\raggedright\arraybackslash}p{0.42\textwidth}@{}}
\toprule
Task & Language instruction & Varying conditions and success criteria \\
\midrule
Lift Cup & Pick up the cup and put it back down the number of times shown on the card. & The requested count ranges from 3 to 10. The robot must complete exactly that many repetitions and return the cup; either too few or too many repetitions is a failure. \\
\addlinespace
Cover Ducks & Cover the ducks with the cups from left to right, then uncover the pink, green, and blue ducks in that order. & Initial positions and color order vary. The robot must complete the covering stage and uncover all three ducks in the pink--green--blue order; any order violation is a failure. \\
\addlinespace
Route Recall & Watch the order in which the blue boxes are pointed at, then place the cup into the boxes in the same order. & A human points to boxes in a varying sequence of 3--6 steps, including possible repeats. Success requires reproducing the entire demonstrated sequence, including repeated visits. \\
\addlinespace
Put Back Cup & Move the cup to the center, return the arm to its home position, then pick up the cup and put it back in its original box. & The initial box varies. Success requires the complete sequence: move the cup to the center, return the arm home, and return the cup to its original box. \\
\bottomrule
\end{tabular}
\end{table}

For each method, we evaluate 10 episodes per task per embodiment, yielding 40 episodes per embodiment and 80 in total. Each task's success rate pools the 20 episodes from both embodiments; the overall score is the unweighted mean of the four task scores. The same task protocols and success criteria apply to all compared methods.

\subsection{Baseline Results and Comparison Settings}
\label{app:baseline_comparisons}

Table~\ref{tab:baseline_result_sources} lists the sources of published simulation baseline scores; these methods are not reimplemented or retrained here. We compute group, category, and overall averages from task-level scores before rounding, except for EventVLA's RMBench overall score, for which we retain the originally reported 67.8\% rather than recomputing it from the displayed task scores. For RoboMME, we select each method's highest-scoring non-oracle variant by overall success, excluding ground-truth symbolic memory. Published scores retain their original evaluation settings: the RoboMME report uses 50 episodes per task and averages over three seeds and three checkpoints; we evaluate our policy on 100 episodes per task.

\begin{table}[!ht]
\centering
\caption{\textbf{Sources of simulation baseline scores.} These are published results, not new baseline runs.}
\label{tab:baseline_result_sources}
\small
\setlength{\tabcolsep}{4pt}
\renewcommand{\arraystretch}{1.15}
\begin{tabular}{@{}>{\raggedright\arraybackslash}p{0.15\textwidth}>{\raggedright\arraybackslash}p{0.30\textwidth}>{\raggedright\arraybackslash}p{0.50\textwidth}@{}}
\toprule
Benchmark & Numerical source & Coverage \\
\midrule
RMBench & EventVLA, Appendix Table 8~\citep{eventvla} & All baseline task scores in Table~\ref{tab:rmbench_results}. DP, ACT, $\pi_{0.5}$, X-VLA, and Mem-0 are also reported in RMBench~\citep{rmbench}. \\
\addlinespace
RoboMME & RoboMME benchmark report~\citep{robomme} & All baseline task scores underlying Table~\ref{tab:robomme_results}, including the memory variants, past-action baseline, and MemER. \\
\bottomrule
\end{tabular}
\end{table}

For the real-world comparison, $\pi_{0.5}$~\citep{pi05} and $\pi_{\mathrm{MEM}}$~\citep{mem} are fine-tuned from their respective pretrained weights on the same collected demonstrations as EvoMem-VLA. All other training and evaluation conditions are held fixed across methods. Each method is trained separately for each embodiment using the data in Appendix~\ref{app:real_world_details} and evaluated under the protocol in Appendix~\ref{app:real_world_evaluation}.

\section{Ablation Details and Qualitative Analysis}
\label{app:ablation_details}

\subsection{Controlled Ablation Configurations}
\label{app:ablation_configurations}

All ablations are conducted on RMBench. Each variant is trained separately with the same training data, initialization, 80,000-step training budget, and evaluation conditions as the full model. All variants retain the same initial and recent visual observations, and keyframe memory is disabled throughout, as in the full RMBench model. Table~\ref{tab:ablation_configurations} specifies the changes, with quantitative results reported in Table~\ref{tab:rmbench_ablation}.

Table~\ref{tab:rmbench_ablation} reports ablations implemented within our QwenOFT-based framework. The variant without delta tokens and the subtask route retains the initial and recent visual observations and predicts actions directly. It achieves 44.3\% under the same training and evaluation protocol as our other ablations. EventVLA results in Table~\ref{tab:rmbench_results} are taken from the published report; the state-only variant in Table~\ref{tab:rmbench_ablation} is our own ablation, not a reproduction of EventVLA.

\begin{table}[!htbp]
\centering
\caption{\textbf{Configurations of the RMBench ablations.} Visual history and training settings are held fixed. Subtask supervision applies only to subtask-route samples.}
\label{tab:ablation_configurations}
\small
\setlength{\tabcolsep}{4pt}
\renewcommand{\arraystretch}{1.15}
\begin{tabular}{@{}>{\raggedright\arraybackslash}p{0.27\textwidth}>{\raggedright\arraybackslash}p{0.29\textwidth}>{\raggedright\arraybackslash}p{0.19\textwidth}>{\raggedright\arraybackslash}p{0.18\textwidth}@{}}
\toprule
Variant & Change representation & Action routing & Subtask language loss \\
\midrule
EvoMem-VLA & Conditional delta tokens & Task-adaptive & Enabled \\
\addlinespace
w/ naive feature difference & Direct feature subtraction & Task-adaptive & Enabled \\
\addlinespace
w/o subtask route & Conditional delta tokens & Direct only & Disabled \\
\addlinespace
w/o delta tokens & None & Task-adaptive & Enabled \\
\addlinespace
w/o delta tokens \& subtask route & None & Direct only & Disabled \\
\bottomrule
\end{tabular}
\end{table}

The variant without delta tokens removes change tokens while preserving visual state memory and task-adaptive routing. The variant without the subtask route retains delta tokens but removes subtask text from the policy input and the corresponding language supervision, predicting actions directly for all tasks. The naive feature difference variant uses the same ordered frame pairs but replaces the learned Delta Encoder with target-minus-source DINOv3 features, followed by projection into the VLM embedding space. It tests whether directly exposing feature changes is sufficient, without learning the conditional change representation. The joint-removal variant removes both delta tokens and subtask conditioning. Together, the full model and the three removal variants form a two-by-two comparison of delta tokenization and subtask conditioning, measuring each component's contribution with and without the other.

\subsection{Qualitative Trajectory Comparison}
\label{app:qualitative_comparison}

Figure~\ref{fig:rmbench_change_trajectory_comparison} compares the full model and the two change-representation ablations under the same RMBench \emph{Press Button} episode initialization. The instruction requires pressing the left button nine times, the middle button once, and then the termination button. EvoMem-VLA completes this sequence. Without delta tokens, the policy follows an incorrect trajectory; with naive feature differences, it repeatedly presses the middle button and fails. The highlighted next actions illustrate the execution differences that complement the aggregate results in Table~\ref{tab:rmbench_ablation}.

\end{document}